# A General End-to-end Diagnosis Framework for Manufacturing Systems

Ye Yuan[1,2,*], Guijun Ma[3], Cheng Cheng[2], Beitong Zhou[2], Huan Zhao[3], Hai-Tao Zhang[1,2], Han Ding[1,3,*]

[1]State Key Lab of Digital Manufacturing Equipment and Technology, Huazhong University of Science and Technology, Wuhan 430074, P.R. China.

[2]School of Artificial Intelligence and Automation, MOE Key Lab of Intelligent Control and Image Processing, Huazhong University of Science and Technology, Wuhan 430074, P.R. China.

[3]School of Mechanical Science and Engineering, Huazhong University of Science and Technology, Wuhan 430074, P.R. China.

## Abstract

**The manufacturing sector is envisioned to be heavily influenced by artificial intelligence-based technologies with the extraordinary increases in computational power and data volumes[1,2]. A central challenge in manufacturing sector lies in the requirement of a general framework to ensure satisfied diagnosis and monitoring performances in different manufacturing applications. Here we propose a general data-driven, end-to-end framework for the monitoring of manufacturing systems. This framework, derived from deep learning techniques, evaluates fused sensory measurements to detect and even predict faults and wearing conditions. This work exploits the predictive power of deep learning to automatically extract hidden degradation features from noisy, time-course data. We have experimented the proposed framework on ten representative datasets drawn from a wide variety of manufacturing applications. Results reveal that the framework performs well in examined benchmark applications and can be applied in diverse contexts, indicating its potential use as a critical corner stone in smart manufacturing.**

## Introduction

In recent decades, it has been envisioned that sensory data measured in manufacturing processes, including vibration, pressure, temperature, and energy data, can be used as features for artificial intelligence (AI) algorithms[3]. AI algorithms have the potential to localize faults or even predict faults before they occur. In this way, run-to-failure maintenance could be replaced by condition-based or predictive maintenance which would be more effective in reducing unnecessary maintenance cost while guaranteeing the reliability of the machinery[4]. However, the existing diagnosis and monitoring techniques most focus on specific tasks, advanced approaches should be developed to form a general framework to produce satisfied performances after simple tuning of parameters in different manufacturing applications.

Model-based and data-driven approaches are two main techniques for diagnosis and monitoring. Model-based approaches for fault monitoring use mathematical models to provide insights into the failure mechanism of mechanical systems[5,6]. Faults are diagnosed by monitoring discrepancies between model predictions and the actual measurements. With the increasing volume of data captured from sensors during manufacturing processes, data-driven approaches have been gaining considerable attention[7,8]. Data-driven approaches are featured by building models without using the knowledge of the failure mechanism, but can perform excellent prediction results[7,8]. The measured sensory signals have often been processed via feature extraction[9] to represent the complete signals manually. The extracted features are then used to train the system using standard classification and regression methods to allow predictions to be made in a case-by-case manner[10-12]. However, both model-based and data-driven approaches are highly tuned to applications, and could not be generalized to other applications without substantial efforts. Consequently, there exists an urgent need for a method that can simultaneously provide convenience for feature extraction and offer universality for use in diverse manufacturing applications.

On the other hand, the convolutional neural network (CNN)[13], as an important type of deep learning, obtained remarkable results in ImageNet in 2012[14] and has gradually become a representative method that is used in medical diagnosis[15], image recognition[16] and speech recognition[17] applications. When compared with other machine learning algorithms, the advantage of CNN is that it enables automatic feature extraction from raw data and can thus eliminate any dependence on prior knowledge[18], which brings inspiration that CNN could provide unified, end-to-end solutions to industrial problems.

This paper transforms manufacturing monitoring problems into a unified supervised learning framework. In particular, it proposes a general end-to-end framework, i.e., a CNN that can extract features automatically and solve the problems accurately. Its outperformance is verified via using ten measurement datasets for different manufacturing problems. Two open benchmark datasets including Case Western Reserve University's bearing data[19] and hydraulic system data[20], five experiment datasets performed in the lab including airplane girder simulation damage data[21], broken tool data, the bearing data[21], tool wear data, and gearbox data[22] were all converted into classification problems. Moreover, National Aeronautics and Space Administration (NASA) tool wearing data[23], battery data[24] and the Center of Advanced Life Cycle Engineering (CALCE) battery data[25] were converted into regression problems. Higher than 95% accuracies are achieved using a unified CNN framework for manufacturing diagnosis problems, meanwhile, small monitoring errors are achieved for condition monitoring problems, indicating the proposed framework has a good application prospect in the manufacturing field. In addition, robustness of the proposed framework is investigated by adding different levels of additive noises to the raw signals in diagnosis tasks.

## Results

**Rolling bearing fault detection and classification** is used here as an illustrative example for the proposed framework, other applications can be found in the Supplementary Information. Rolling bearings are vital components in many types of rotating machinery, ranging from simple electrical fans to complex machine tools. More than half of machinery defects are generally related to bearing faults[26]. Typically, a rolling bearing fault can lead to machine shutdown, chain damage, and even human casualties[26]. Bearing vibration fault signals are usually caused by localized defects in three components: the rolling elements; the outer race, and the inner race. When bearings degrade near the end of their lifetimes, instances of deformation, cracking, and burning among these components may cause spindle deviation and further serious damage to mechanical systems.

A bearing data set provided by the Case Western Reserve University (CWRU) data center (shown in Figure 1(b))[19]-- which is regarded as a benchmark for the bearing fault diagnosis problem-- is used to validate the effectiveness of our proposed framework. An experimental platform (illustrated in Figure 1(c)) was used to conduct the signals to be used for defect detection on bearings with three different fault diameters (7 mils, 14 mils, and 21 mils (1 mil=0.001 inches)). Vibration signals in different conditions from the inner race, the outer race, and the rolling elements for all fault diameters were acquired using accelerometers. The dataset originally consisted of four rotating speeds (1797rpm, 1772rpm, 1750rpm, 1730rpm), and totally had 4 normal samples and 52 faulty samples. We formulate this as a fault diagnosis problem by classifying the fault types as representations of the following three problems: a) binary classification (normal plus faulty conditions), (b) four-way classification (normal plus three main faulty conditions), and (c) ten-way classification (normal plus three main faulty conditions for each of the faulty diameters).

Each sample contains a different number of time-course measurements. To increase the number of samples for training a more accurate model, we reshape the samples here to ensure that each sample had 6000 time-course measurements consistently. In total, 1320 samples are reconstructed from the original dataset. Considering that the potential time dependency existed among the reconstructed samples, we apply three standard cross-validation methods (Random Subsets, Contiguous Block, Independent Sequence[27], shown in Figure 1(e)) to evaluate the performance of the CNN method.  For Random Subsets Method, the entire pre-processed dataset is constructed and then randomly divided into 90% for training (1188 samples) and 10% for validation (132 samples). Figure 1(a) presents the t-SNE visualization[28] of the binary classification features before the final classifier. Features for normal and faulty signals are clearly separated into two clusters indicating that a good classification can be easily obtained by selecting a proper final classifier. Figure 1(c) demonstrates that the classification accuracy will be improved with the increasing of the sample number. We also reveal that more samples are required to obtain a promising result for a more complex problem, such as the ten-way classification task that requires at least 400 samples for training a model with 90% accuracy. Classification results and evaluation metrics are summarized in Figure 2, where all three models achieve 100% (i.e., 132 of 132 validation samples) fault classification, and consistent over different randomization. For Contiguous Block Method, we divide the 1320 samples into training set and test set according to time, and the proportion of the test set varied from 10% to 50% of entire sample number. For Independent Sequence Method, we eliminate the confused dependency issue by dividing the dataset into completely independent training set and test set, e.g. the training set and test set are obtained based on rotating speeds, where random three rotating speeds are used for training and the other one is used for test. Similar results are obtained: 100% (340/340), 100% (340/340), and 98.82% (336/400) for two-way classification, four-way classification, and ten-way classification, respectively. The experiment results of three cross-validation methods are shown in **错误!未找到引用源。**.

Table 1: Three cross-validation methods (Random Subsets, Contiguous Block, Independent Sequence) are employed to verify the effectiveness of the model in CWRU data. 10%, 20%, 30%, 40%, and 50% test scheme for Contiguous Block Method is used. X% means the first 100- X percent of the total data set is used for training and the rest (X percent of the data) is used for test.

| | **Random Subsets** | **Independent Sequence** | **Contiguous Block** | | | | |
|---|---|---|---|---|---|---|---|
| | | | 10% test | 20% test | 30% test | 40% test | 50% test |
| **Two-way** | 100% (132/132) | 100% (340/340) | 100% (132/132) | 100% (262/264) | 100% (396/396) | 100% (528/528) | 100% (660/660) |
| **Four-way** | 100% (132/132) | 100% (340/340) | 100% (132/132) | 99.24% (262/264) | 97.22% (385/396) | 95.83% (506/528) | 96.21% (635/660) |
| **Ten-way** | 100% (132/132) | 98.82% (336/340) | 99.24% (131/132) | 98.86% (261/264) | 98.48% (390/396) | 96.21% (508/528) | 96.67% (638/660) |

For further evaluation of the classification results, we used the following three assessment metrics that are commonly used in machine learning to evaluate the classification performance with validation data using Random Subsets Method: (a) precision, (b) recall, and (c) accuracy, which are defined as follows[29-31]:

$$\text{precision} = \frac{\text{TP}}{\text{TP+FP}} \times 100\%, \tag{1}$$

$$\text{recall} = \frac{\text{TP}}{\text{TP+FN}} \times 100\%, \tag{2}$$

$$\text{accuracy} = \frac{\text{TP+TN}}{\text{TP+TN+FP+FN}} \times 100\%, \tag{3}$$

where the abbreviations TP, FP, FN, and TN denote the numbers of true positives, false positives, false negatives, and true negatives, respectively. In our four-way and ten-way classification, we regarded the first class as the positive class while others are negative classes for computing these metrics. Across the three classification tests, the defined assessment metrics all achieved results of 100%.

These results demonstrate that, without prior knowledge (manufacturing parameters and failure mechanism), measurement data as well as their labels suffice to classify fault types accurately and thereby pinpoint faults' location, which makes the fixing process efficient. In addition, the proposed framework requires average five minutes for training using a standard GTX 1080 GPU. With the obtained trained model, the CNN performs fault-prediction results within 0.05 s on the same GPU, which is fast enough compared to the sampling time of 0.1 s. Therefore, the proposed algorithm can be implemented online to localize faults in real time.

**Generalizability of the proposed CNN framework:** A major feature of the proposed framework can also be generalized for a wide range of other applications with high metrics, including greater accuracy, precision and recall (summarized in Figure 2). Here we focus on two other representative applications:

1. Hydraulic system condition classification[20]: With its excellent performance in creating movement or repetition[32,33], hydraulic system-based equipment have been widely used in many applications including manufacturing, robotics, and steel processing. However, fluid in hydraulic system is highly pressurized, extremely hot and even toxic, which bring high level of hazards to the workers and the surrounding environment. Our CNN fault prediction algorithm for hydraulic system can generate hazard warning signals to prevent chemical burns to the workers, igniting nearby materials, and causing explosions in real time. Hydraulic system condition monitoring is a classification task. We chose CNN as the base model to make predictions for different conditions. Four condition classifications corresponding to different hazard types and levels were conducted: 1) a three-way classification for cooler condition, 2) a four-way classification for valve condition, 3) a three-way classification model for internal pump leakage, and 4) a four-way classification model for hydraulic accumulator. The algorithm achieved accuracies of 100% in both cooler condition and valve condition classifications. Meanwhile, the pump leakage and hydraulic accumulator classifications also achieved satisfactory accuracies, at 98.19% and 99.10%, respectively.

2. NASA lithium-ion battery data for state of health (SOH) estimation[24]: Lithium-ion batteries (LiBs) are the auxiliary or main power sources for many electronic systems, including medical devices, aerospace systems, smart phones, and electric vehicles[34]. Estimating SOH is the key issue to evaluate the LiBs health status. Exceeding the minimum level of the SOH will lead to a rapid performance degradation and even affect system safety. A benchmark of industrial lithium-ion battery data obtained by NASA is used to estimate battery SOH. CNN models are trained for this dataset and the smallest average RMSE value 0.0172 mm is achieved with respect to smallest error[35] of 0.0264 mm that has been achieved in previous related work. Detailed descriptions of the data structures and the established models for these applications and several further, diverse, cases can be found in the Supplementary Information.

**Interpretability of the proposed CNN framework:** To validate the generality of the proposed CNN algorithm framework for fault-prediction, it is key to understand how CNN exacts meaningful features from manufacturing data. However, interpreting deep neural networks remains a notoriously difficult task in the literature. Inspired by practical successful studies in medicine[36] and biology[37], we have developed a method for manufacturing data to visualize the general features, such as frequency, phase or amplitude, extracted by CNN model that contribute to the fault-prediction. Note that, in this study, we only focus on revealing the relationship between the convolutional layers (outputs before fully-connected layers) and the hidden features in manufacturing data of our study.

Time-series signal $(F_n,\ \phi_n, a_n,\ u)$ as the most common form in manufacturing data is constructed as the sum of harmonically related sinusoids and expressed by Fourier series form **(Materials and Methods)**, based on [ISO 2041:2018][38], with varying frequencies $F_n$ (associated with fault frequencies, noise frequencies, and resonance frequencies of mechanical components), phases $\phi_n \in [0, 2\pi]$, amplitudes $a_n$ (associated with damage levels on mechanical components), or/and white noise $u$ (associated with environmental noises). Binary classification experiments (**Materials and Methods**), with class A signals ($F_n^A, \phi_n^A, a_n^B, u$) and class B signals ($F_n^B, \phi_n^B, a_n^B, u$), used to visualize the contribution of the convolutional layers are conducted. Several effects with respect to different frequencies, phases, amplitudes, and noises are shown in Figure 4.

Starting from the basic single sinusoid function (class A signal), the fault signals (class B signals) are with varying features corresponding to different frequencies, phases, or amplitudes from the left to the right plot respectively (Figure 4(a)). Frequency domain results in the first plot reveal that the extracted features (feature A and feature B) from convolutional layers are with the same frequencies as the input signals (class A and class B). This is a clue to classify manufacturing data with different frequency components due to wearing, broken, or deformation. Polar coordinate result in the second plot appears an interesting phenomenon that the phase difference between two features extracted is around $\frac{\pi}{2}$, which is equal to the initial phase difference. The third plot show the CNN ability to distinguish amplitude difference of manufacturing data. Fault signals (class B signals) are with five times amplitude than the normal signal (class A signal), results demonstrate that the magnitudes of features after convolutional layers for normal and fault signals present the same proportional relationship as the initial input signals. In Figure 4(b), with an additional Gaussian noise compared to Figure 4(a), results revealed that the CNN can ignore the redundant noise and extract the valuable information (almost the same features as Figure 4(a)) for classification. Figure 4(c) shows a more complex case which is a combination of two sinusoid signals, features of fault signals after the convolutional operations (class B signals) can be also easily distinguished in both frequency domain and polar coordinate.

The final fault prediction decision obtained by the CNN is a collective impact of all the above discussed coefficients with respect to signal frequencies, phases, amplitudes, and biases. In this study, we attempt to give a plausible interpretation of the CNN framework for manufacturing data, from simple to complex cases. We reveal that CNN is successful for capturing the features of manufacturing problem due to the fact that the time-series signals (the most common form of manufacturing data) are compositional hierarchies.

**Robustness of the proposed CNN framework:** To test the robustness of the proposed method, additive noise is added to each sample. For Case Western Reserve University's bearing data, added noise whose power varies from 0% to 500% of each original signal power is used to examine the robustness of the CNN framework, the prediction results for three classification tasks are shown in Figure 1(d). In addition, the intensity of added noise in other diagnosis applications are listed in Table 2. The classification applications still get high accuracies when the range of additive noise within a certain range, which reveals the proposed CNN framework have a satisfactory robustness.

Table 2: The range of added noise in each manufacturing application.

| Data | Noise range |
|---|---|
| CWRU bearing data | 0% - 500% |
| Broken tool data | 0% - 600% |
| Bearing data collected by our lab | 0% - 200% |
| Airplane girder simulation damage data | 0% -20% |
| Process data | 0% - 150% |
| Gearbox data | 0% - 500% |
| Hydraulic system data | 0% -1% |

**Discussions**

In summary, we have demonstrated the effectiveness of proposed framework for usage in manufacturing systems. Using a unified framework, we have tested the proposed deep learning algorithm against a large number of critical diagnostic tasks in a variety of applications. The proposed end-to-end framework achieves the highest accuracy reported for both the benchmark datasets and our own datasets. The interpretation ability of fault-prediction in CNN model provides valuable information for understanding why deep learning can make diagnosis decision on manufacturing data with different frequencies, amplitude, and phases. With the designed hardware implementation specified in the Supplementary Information, our proposed algorithm framework could be easily applied to dataset in other industry applications.

This framework entails some limitations, which shall be listed as our future work. The limitations include: First of all, the method has a few hyper parameters to tune, in order to achieve best performance, cross validation or optimization algorithm could be used to select the hyper parameters, such as kernel size, the number of strides and the number of layers. Secondly, given the number of parameters in the constructed model, it requires a large amount of data to train, this may not be feasible for certain applications.

## Materials and Methods

### Datasets

The datasets used in this manuscript are of the following types: open-accessible data, competition data, experimental data collected in our lab, and real production data provided by industrial partners with permission. These datasets are composed of sensory current signals, force signals, vibration signals or acoustic emission signals, or their combinations, which are processed for the classification or regression tasks.

### Main idea

We convert practical problems into supervised classification and regression tasks and solve them using deep learning technique. An end-to-end algorithm is proposed to automatically discover the hidden features needed for learning and prediction without prior knowledge. We develop a novel framework based on CNN that performs fault diagnosis and prediction and regression based on the raw data. The framework of the proposed CNN is shown in Figure 3, in which we construct a fully automated closed-loop system: a CNN model is fed with the sensory measurements and automatically extracts the features for classification or prediction. The results learned by the CNN are then fed back to the machine for decision making, for instance, whether a maintenance action is required.

### Pre-processing

We normalize the measurements in each dataset in several ways as detailed in the Supplementary Information. More specifically, for datasets with a small number of time-course measurements, such as CWRU bearing dataset, we divided the total features to a constant length in each sample without affecting the periodicity of the data. For prediction tasks, such as the Case 8, dataset is transformed by a standardisation as specified in the Supplementary Information.

Parameter-tuning

We propose to fine-tune the CNN model according to different classification and prediction objectives, with a fixed max-pooling size of $1 \times 2$. To extract fewer features, stride sizes (i.e., the sliding window size) in CNN models are set to be, for example, 500 or 1000 in a data sequence with tens of thousands of dimensions, 100 or 200 in a data sequence with thousands of dimensions. The basic components of the proposed CNN model are stacked with input data, CNN layers, and a fully connected layer (including an output layer). For classification problems, the number of nodes N in the output layer are equal to the number of fault types. For regression problems, N is set to one. For detailed model parameters of difference applications, please refer to Supplementary Information.

Convolutional neural networks

CNNs consist of convolutional layers, Max-pooling layers, a flatten layer and fully connected layers with a final N-way prediction layer. In essence, the CNN uses raw data $\mathrm{I} \in \mathbb{R}^{\mathrm{k}}$ as inputs and outputs are classification or regression results $\hat{y}$, i.e.,

$$\hat{y} = act(FCN(\mathrm{Flatt}(pool(ReLU(conv(I)))))). \tag{4}$$

The convolutional layer (*conv*) uses a number of filters to discretely convolve with the input data. We define a weight vector $H \in \mathbb{R}^{\mathrm{m}}$, a data vector $I \in \mathbb{R}^{\mathrm{k}}$ computed from raw data, and a constant $b$ of a bias. In a convolutional process, stride is the distance between two sub-convolution windows, and we define it as parameter $d$. We define the $\mathrm{i}_{\mathrm{th}}$ sub-vector of $I$, i.e., $I^{(i)} = \left[I^{1+(i-1)d}, I^{2+(i-1)d}, \ldots, I^{m+(i-1)d}\right]^T$ $\left( i = 1,2, \ldots, \frac{k-m}{d} + 1\right)$. The idea of a one-dimensional convolution is to take the product between the vector $H$ and the sub-vector $I^{(i)}$ of raw data, which reads as follows:

$$S^{(\mathrm{i})} = I^{(\mathrm{i})} * H + b = \sum_{j=1}^{m} I^{j+(i-1)d} H^j + b \,, \tag{5}$$

where $H^j$ is the $j_{th}$ element of vector $H$, $j=1,2,\ldots,m$. When conducting a convolutional process, the number of filters (different filters have different initial vector *H*) are set to determine the depth of the convolutional results. Since the process of convolution between each filter and data uses weight sharing, the number of training parameters and complexity of the model are greatly reduced. As a result, computational efficiency is improved.

An activation function named Rectified Linear Unit (*ReLU*) is followed by each convolutional layer, which has the following form:

$$U^i = ReLU\left(S^{(i)}\right) \triangleq max\left(0, S^{(i)}\right) . \tag{6}$$

ReLU avoids gradient vanishing with respect to other functions when optimizer calculates gradient descent, meanwhile guarantees the sparsity in convolutional networks, which significantly reduces the training time compared with other activation functions. Above operations lead to the results of $U = [U^1, \dots, U^i, \dots, U^{\frac{k-m}{d}+1}]^T$.

Then Max pooling (*pool*) is chosen here:

$$pool(U^i) := max_{\ell=1}^{p} U^{\ell+(i-1)e} \ , \ \forall i = 1,2, \dots, \frac{k-m}{d} + 1, \tag{7}$$

where *p* is the pooling size, and *e* is the stride size.

After convolution and pooling, the data is fed into a flatten layer (*Flatt*), data is transformed into a one-dimensional structure in *Flatt,* denoted as $F = \left[F_1, F_2, \dots, F_q\right], q$ is the length of data after the flatten layer to facilitate data processing in the fully connected layers (*FCNs*). Then *FCNs* combined with *ReLU* activation function is utilized to realize dimensionality reduction, which can be written as:

$$O = ReLU(W \cdot F), \tag{8}$$

where *W* are the weights of FCNs, $O = [O_1, O_2, \dots, O_N]$ is the output of the *FCNs*, and '·' is dot product. N is the number of faulty types in classification task and N=1 in regression task.

The output activation function (*act*) uses softmax function for classification problem or sigmoid function for regression problem. For classification, the estimated result $\hat{y} = act(O)$ can be shown as:

$$\hat{y}_n = \frac{e^{O_n}}{\Sigma_{j=1}^{N} e^{O_j}} \ , \qquad \text{for n} = 1,2, \dots \text{N}. \tag{9}$$

And for regression, $\hat{y} = act(O)$ is:

$$\hat{y} = \frac{1}{1+e^{-O}} \,. \qquad (10)$$

In the training process, to minimize the difference between the predicted scores and the ground labels in the training data, cross-entropy $L_{ce}$ and least-squares $L_{ls}$ are chosen as the loss function for classification problem and regression problem respectively, which are defined in Equations (11) and (12):

$$L_{ce} = -\frac{1}{q}\Sigma_{i=1}^{q}\Sigma_{n=1}^{N}1\{y^{(i)} = n\}\, log\, \hat{y}^{(i)} + \left(1 - 1\{y^{(i)} = n\}\right) log\left(1 - \hat{y}^{(i)}\right), \qquad (11)$$

$$L_{ls} = \frac{1}{q}\Sigma_{i=1}^{q}\left(y^{(i)} - \hat{y}^{(i)}\right)^{2}. \qquad (12)$$

where $y^{(i)}$ is the real output of the $i_{th}$ training measurement and $q$ the total number of training measurement. The term $1\{y^{(i)} = n\}$ in Equation (10) is the logical expression that always returns either zeros or ones.

Once the loss function has been chosen, we use standard optimizers such as stochastic gradient descent (*SGD*)[39] or *Adam*[40] for parameter training in back-propagation to update weights. The final CNN model weights refresh until the predefined maximum iteration to yield a lower loss.

Interpretation of CNN model for manufacturing data

In order to interpret how the CNN model learns from manufacturing data, we consider a time-series signal with the most common form of manufacturing data, which is modelled as the sum of harmonically related sinusoidal functions:

$$v^{(i)} = \sum_{n=1}^{N} a_n \sin(2\pi F_n x^{(i)} + \phi_n) + u^{(i)}\,,\ \text{for } x^{(i)} \in [0, 0.4] \text{ and } i = 0, 1, \dots, 4095 \qquad (13)$$

where $v^{(i)}$ is the magnitude of $i$-th measurements and $u$ is the Gaussian noise. Features of the time-series signal $v^{(i)}$ corresponds to four coefficients: the sinusoid frequencies $F_n$, the amplitudes $a_n$, the phases $\phi_n \in [0, 2\pi]$, and the Gaussian noise $u$, respectively.

To provide a clear interpretation of convolutional layers, we conducted binary classification (class A and class B) experiments by giving different numbers to one of four coefficients, while keep the other three coefficients unchanged (Figure 4). For each binary classification task, we duplicated 100 times of class A signal $v_A = \left[v_A^{(1)}, \dots, v_A^{(i)}, \dots, v_A^{(4095)}\right]^T$ and class B signal $v_B =$ $\left[v_B^{(1)}, \ldots, v_B^{(i)}, \ldots, v_B^{(4095)}\right]^T$ as the samples for training and validation. Randomly, 90% samples are used for model training and the other 10% samples validation, and 100% accuracy is obtained. Class A and B signals are processed to convolutional operation using Equation (4-7) to obtain the features for visualization. Figure 4 analyses the extracted features in frequency domain or polar coordinate corresponding to different coefficients in Equation (13).

Cross validation:

Random Subsets approach is used for dividing the training set and test set in all the classification tasks. For the datasets who have small sample sizes, such as tool broken data, blades processing data and gearbox data, we randomly split the dataset into 80% (training), and 20% (validation). For other classification tasks with large sample sizes, we randomly split the dataset into 90% for training and 10% for validation. For CWRU data (Case 1), bearing data (Case 3), and gearbox data (Case 6), two other cross-validation methods (Contiguous Block, Independent Sequence) are used to verify the effectiveness of CNN. Contiguous Block Method utilizes the latter part of samples which are reconstructed from one long time-series as test set while the other part as training part. 10%, 20%, 30%, 40%, and 50% test scheme for Contiguous Block Method is used. X% means the first 100- X percent of the total data set is used for training and the rest (X percent of the data) is used for test. Independent Sequence method divides the training set and test set as independent time-series. For three prediction tasks, leave-one-out cross validation is used. For instance, NASA battery data has three degradation data, and we randomly use two for training and one for validation each time, then three models are trained to validate three test data, respectively.

Robustness analysis:

In order to verify the robustness of the proposed method in each classification task, additive noise, whose power is a variational percentage (expressed as S%) of each original signal power, is added to each sample (see in Equation (14)). Similar cross validation approaches are used in each case. Figure in which the accuracy varies with power of the noise is given in each case.

$$\text{noise power} = \frac{1}{k} \cdot S\% \cdot \sum_{j=1}^{k} (I^j)^2 \tag{14}$$

where $I$ is the vector of raw data, $j$ is sampling point number, and $k$ is the length the raw data.

**Data availability statement**

The bearing fault, aircraft girder, and aero engine blade processing datasets can be downloaded at Manufacturing Network Platform that we built: http://mad-net.org:8085/. The CRWU bearing dataset is available at http://www.eecs.cwru.edu/laboratory/bearing. The NASA tool wear dataset can be downloaded from https://ti.arc.nasa.gov/tech/dash/groups/pcoe/prognostic-data-repository/. NASA and CALCE battery datasets are available at http://ti.arc.nasa.gov/project/prognostic-data-repository and https://web.calce.umd.edu/batteries/data.htm#, respectively. The hydraulic system dataset is available at https://archive.ics.uci.edu/ml/datasets/Condition+monitoring+of+hydraulic+systems#. The experimental data including gearbox, and tool broken datasets are available from the corresponding author upon request.

**Acknowledgements** This work was supported by National Natural Science Foundation of China through projects 91748112, 5135004. We thank Prof. Bin Li, Dr. Bo Luo, Prof. Tao Zhou, Prof. Sijie Yan for early insightful discussion. We thank Prof. Shenfang Yuan, Prof. Jihong Chen, Prof.

Limin Zhu, Prof. Ke Li for kindly sharing their valuable data. We thank Mr. Anthony Haynes, and Mr. David MacDonald from Liwen Bianji, Edanz Editing China (www.liwenbianji.cn/ac), for editing the English text of a draft of this manuscript.

**Author Contributions** Y.Y. conceptualized the algorithm. G.M. developed the algorithms under the supervision of H.D. and Y.Y.. H.D. G.M. C.C., B. Z. and Y.Y. designed and analyzed the experimental data. All authors designed and discussed the study and wrote the paper.

**Author Information** The authors declare no competing financial interests. Readers are welcome to comment on the online version of the paper. All data and computer code needed to evaluate the conclusions in the paper are available from the corresponding author upon request. Correspondence and requests for materials should be addressed to Y.Y. (yye@hust.edu.cn) and H.D. (dinghan@hust.edu.cn).

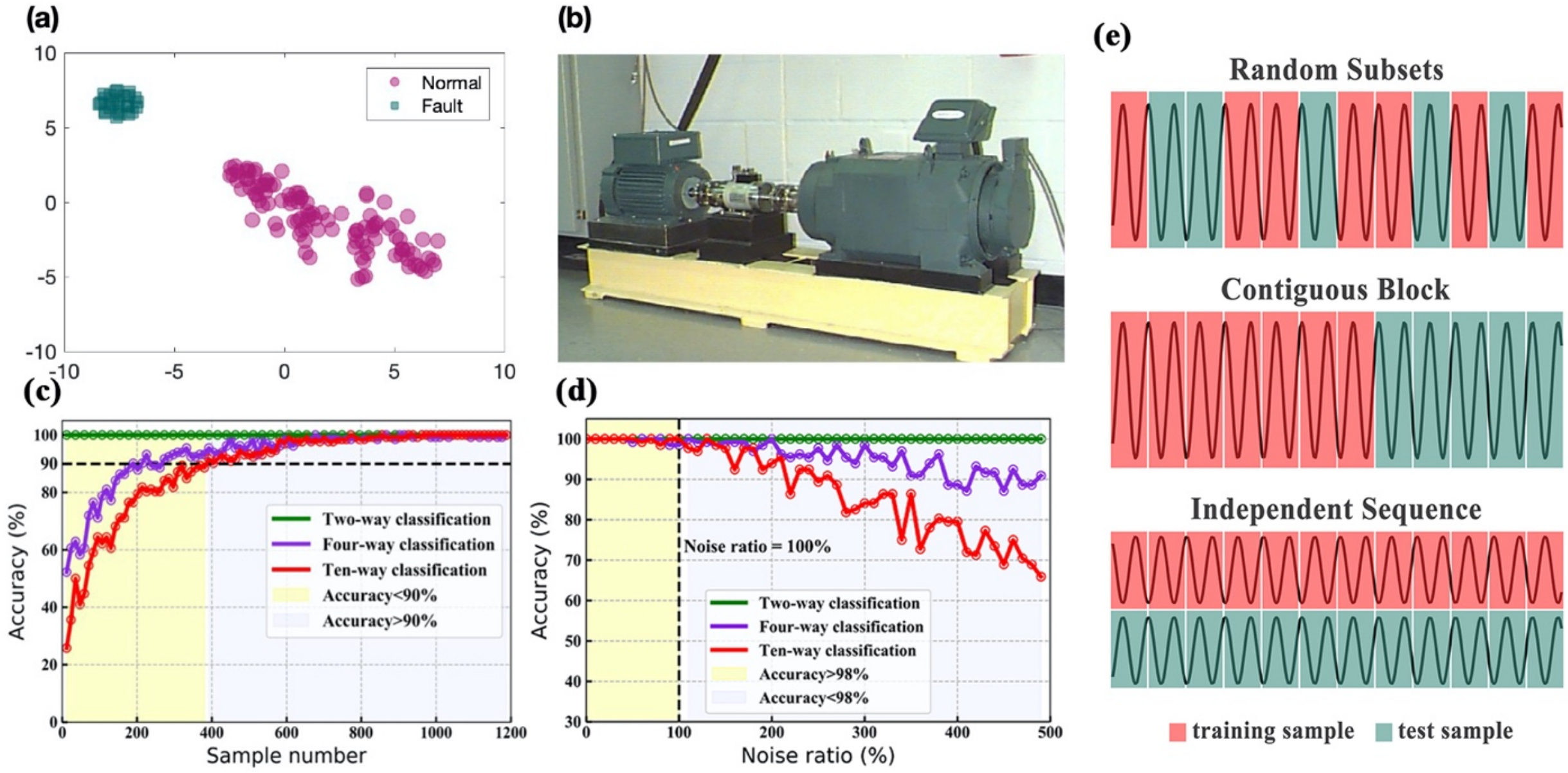

Fig. 1: Classification results on the CWRU bearing test dataset, visualized in a 2D feature space. **(a)**. t-SNE visualization of the binary classification task (normal and fault), **(b)**. The experiment platform in CWRU bearing data center[41], **(c)**. evolution curves of accuracy variations regarding two-way (binary), four-way, and ten-way classification tasks, where sample number is increased from 10 to 1200, **(d)**. the curves of accuracies with noise ratio varies from 0% to 500% for three classification model, where the accuracies all surpass 98% when the noise ratio is less than or equal to 100%, **(e)**. schematic diagram of three cross-validation methods (Random Subsets, Contiguous Block, and Independent Sequence).

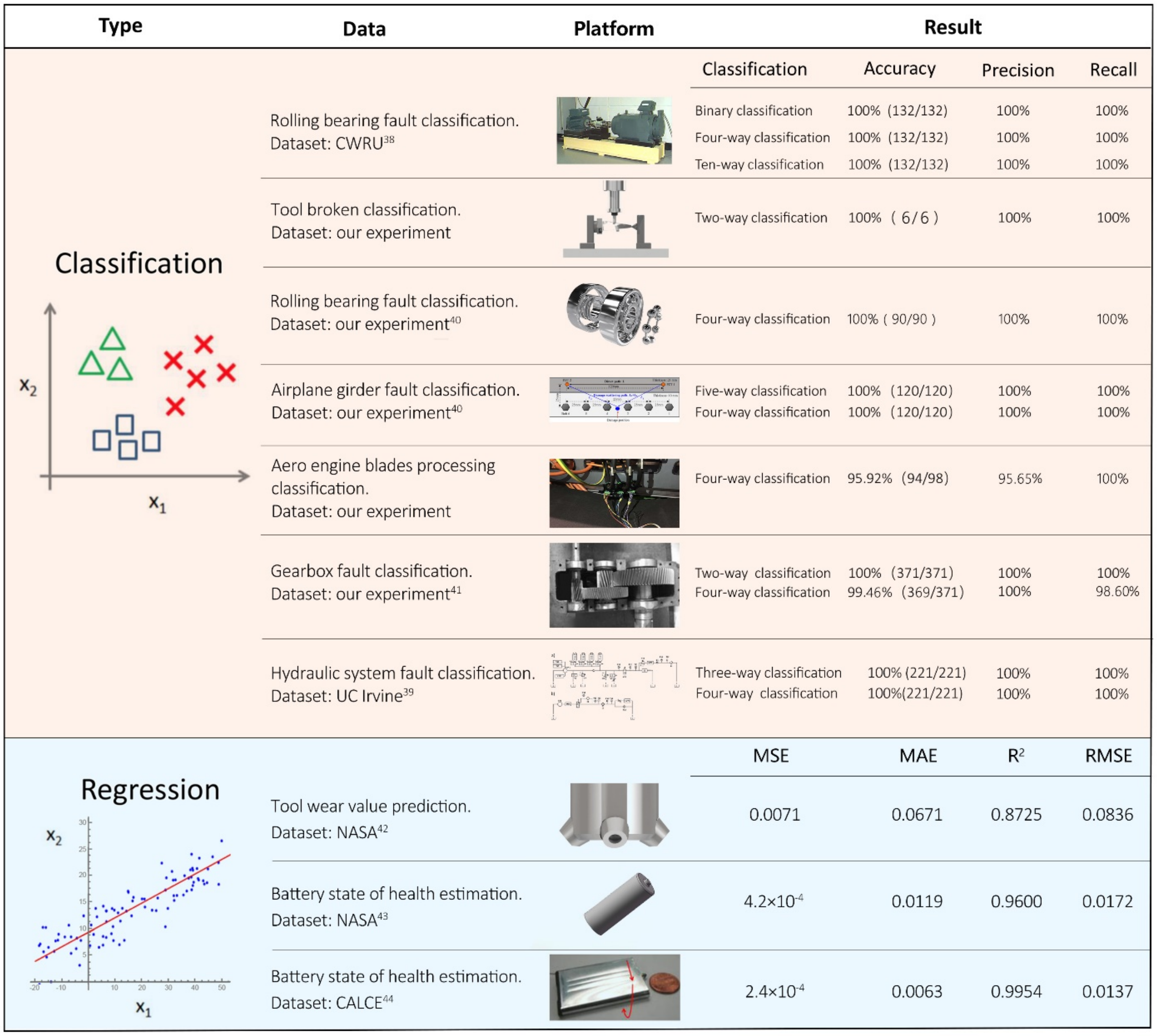

| Type | Data | Platform | Classification | Accuracy | Precision | Recall |
|---|---|---|---|---|---|---|
| Classification | Rolling bearing fault classification. Dataset: CWRU[38] | | Binary classification | 100% (132/132) | 100% | 100% |
| | | | Four-way classification | 100% (132/132) | 100% | 100% |
| | | | Ten-way classification | 100% (132/132) | 100% | 100% |
| | Tool broken classification. Dataset: our experiment | | Two-way classification | 100% ( 6/6 ) | 100% | 100% |
| | Rolling bearing fault classification. Dataset: our experiment[40] | | Four-way classification | 100% ( 90/90 ) | 100% | 100% |
| | Airplane girder fault classification. Dataset: our experiment[40] | | Five-way classification | 100% (120/120) | 100% | 100% |
| | | | Four-way classification | 100% (120/120) | 100% | 100% |
| | Aero engine blades processing classification. Dataset: our experiment | | Four-way classification | 95.92% (94/98) | 95.65% | 100% |
| | Gearbox fault classification. Dataset: our experiment[41] | | Two-way classification | 100% (371/371) | 100% | 100% |
| | | | Four-way classification | 99.46% (369/371) | 100% | 98.60% |
| | Hydraulic system fault classification. Dataset: UC Irvine[39] | | Three-way classification | 100% (221/221) | 100% | 100% |
| | | | Four-way classification | 100%(221/221) | 100% | 100% |

| Type | Data | Platform | MSE | MAE | $R^2$ | RMSE |
|---|---|---|---|---|---|---|
| Regression | Tool wear value prediction. Dataset: NASA[42] | | 0.0071 | 0.0671 | 0.8725 | 0.0836 |
| | Battery state of health estimation. Dataset: NASA[43] | | $4.2\times10^{-4}$ | 0.0119 | 0.9600 | 0.0172 |
| | Battery state of health estimation. Dataset: CALCE[44] | | $2.4\times10^{-4}$ | 0.0063 | 0.9954 | 0.0137 |

Fig. 2: Summary of classification and regression results of different datasets. The datasets for classification problems include: CWRU bearing data; tool broken data; bearing data; airplane girder data; blades processing data; gearbox data; and hydraulic system data. The datasets for supervised regression problems include: NASA tool wear data; NASA battery data; and the CALCE data. Specifically, for the multi-classification problem, we define the first class as the positive class to calculate the precision, recall and accuracy according to Equations (1), (2) and (3).

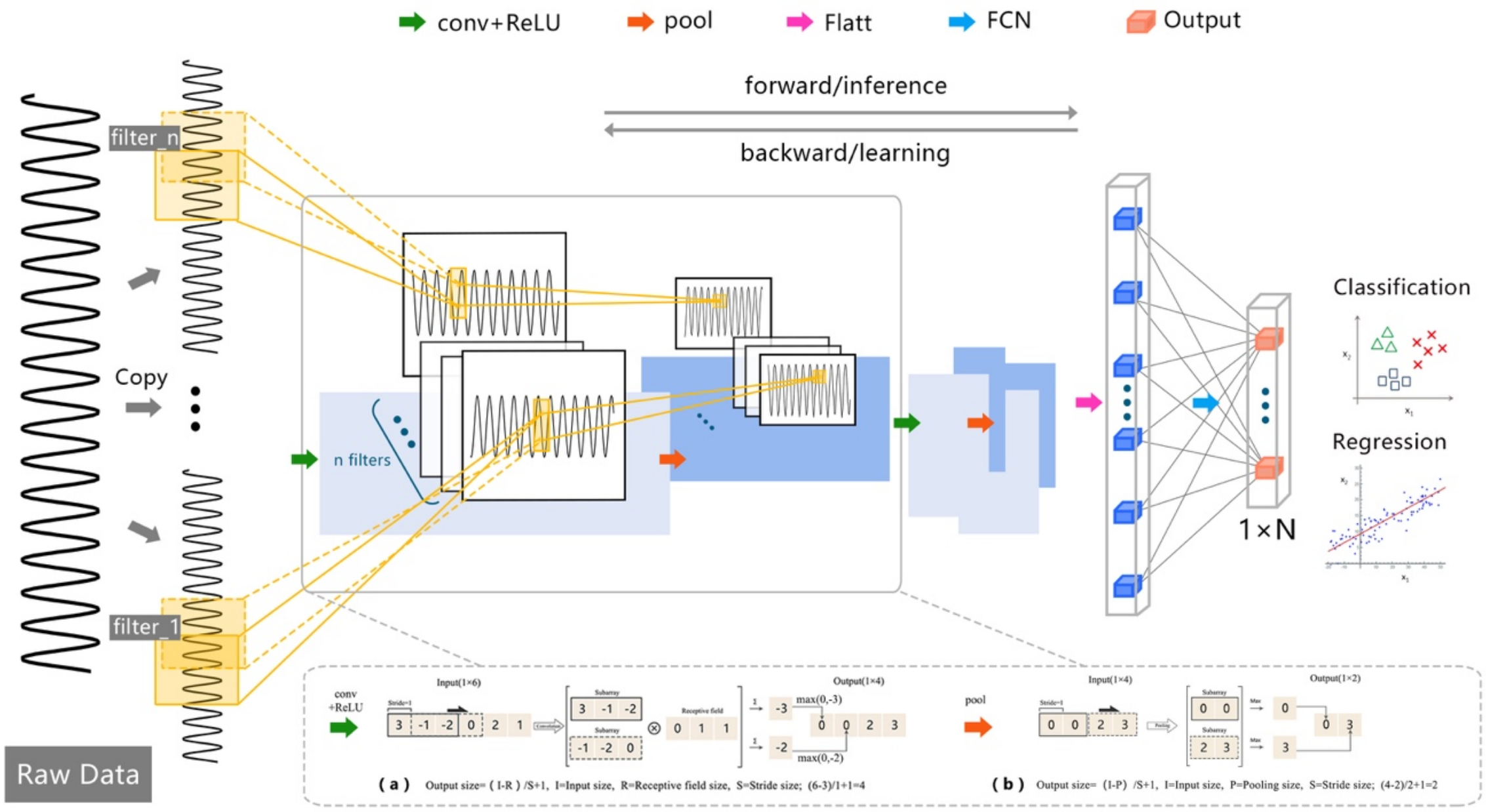

Fig. 3: An illustration of a CNN model for a classification/regression task. This framework is a fully-automated system. Raw data is generated by the manufacturing system and processed and it goes through the CNN operations. The input raw data is passed through convolutional layers **(a)**, max pooling layers **(b)**, and fully connected layers as explained in the **Materials and Methods**. A flatten operation is employed before the data fed into the first fully connected layer. The output layer with $1\times N$ size result (N is an integer for classification categorizes or equals to 1 for regression) from the CNN model can be fed back to the manufacturing system for decision making.

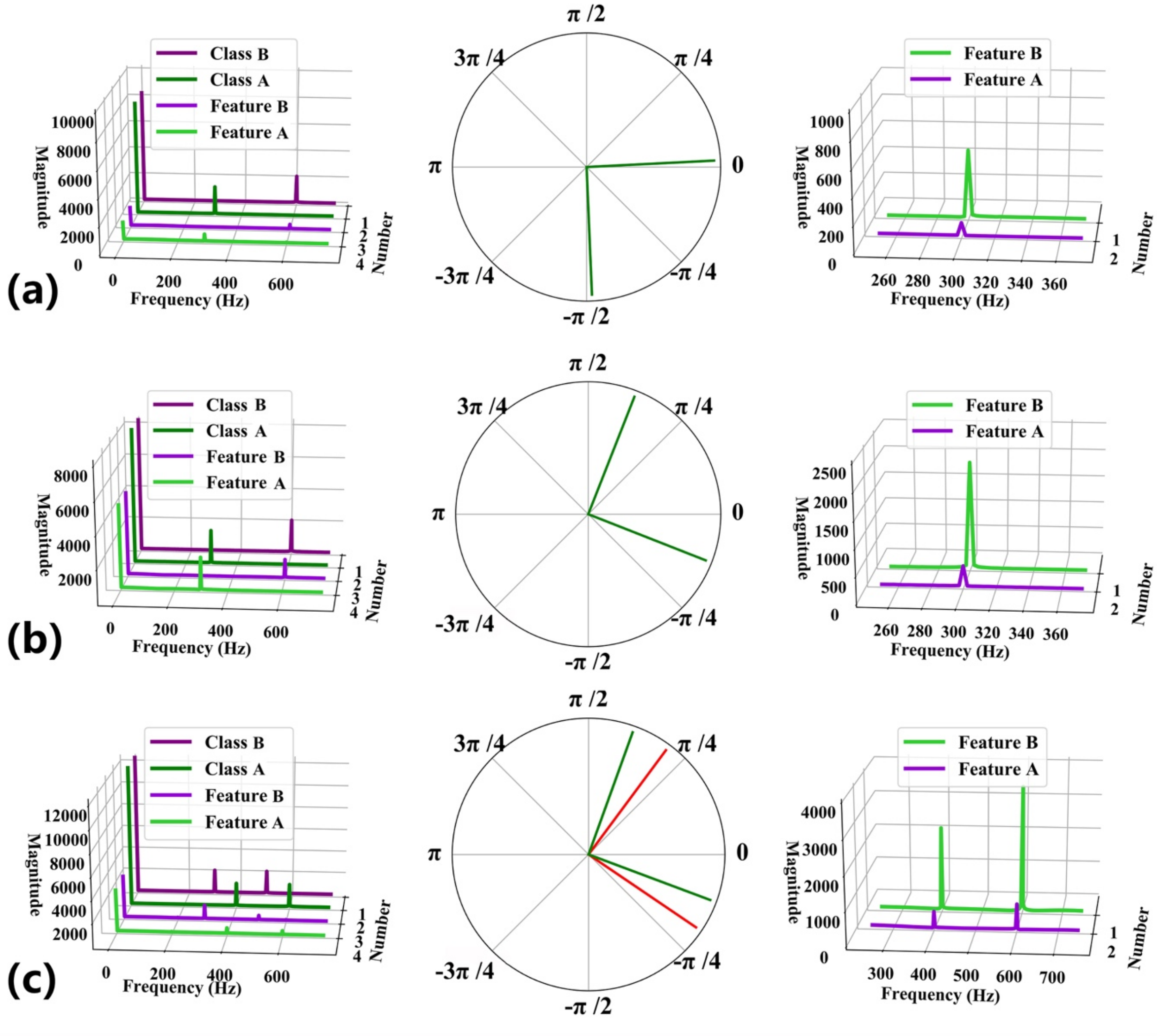

Fig. 4 : Interpretation of CNN model for manufacturing data. Two classes (class A and class B) of time-series signals with different coefficient settings ($F_n$, $\phi_n, a_n, and\ u$ in Equation (12)) are formulated to represent the manufacturing data. The output features (feature A and feature B) of convolutional layers through Equation (4-7) are presented in frequency domain (first and third columns) and polar coordinate (second column). (a). Single sinusoid function case. Class A signal is with fixed coefficients without noise (u=0), where $F_1^A = 300Hz$, $\phi_1^A = 0$, and $a_1^A = 1$. Meanwhile, coefficients of class B signal are changed sequentially from the left plot to the right plot, the rest two coefficients remain the same as class A signal, where in the first plot $F_1^B = 600Hz$, in the second plot $\phi_1^B = \frac{\pi}{2}$, and in the third plot $a_1^B = 5$. (b). Same coefficient settings of case (a), plus an additional Gaussian noise u where $u \sim N(0, 0.05)$ for all class A and B signals.

(c). Sum of two sinusoid functions case. Class A signal is with fixed coefficients without noise (u=0), where $F_1^A = 400Hz$, $F_2^A = 600Hz$, $\phi_1^A = 0$, $\phi_2^A = 0$, $a_1^A = 1$, and $a_2^A = 1$. Same as (a), coefficients of class B signal are changed sequentially from the left plot to the right plot, where in the first plot $F_1^B = 300Hz$ and $F_2^B = 500Hz$, in the second plot $\phi_1^B = \frac{\pi}{2}$ and $\phi_2^B = \frac{\pi}{2}$, and in the third plot $a_1^B = 5$, $a_1^B = 5$.